\documentclass[sigconf]{acmart}

\usepackage{amsmath,amsfonts}

\usepackage{amssymb}
\usepackage{booktabs}
\usepackage{array}
\usepackage{multirow}
\usepackage{tikz}
\usepackage{fontawesome5}
\usetikzlibrary{positioning,arrows.meta,fit,backgrounds,calc,decorations.pathreplacing}
\usepackage{subcaption}
\setcopyright{none}
\renewcommand\footnotetextcopyrightpermission[1]{}
\newcommand{\omniBase}{\texttt{jina-\allowbreak embeddings-\allowbreak v5-\allowbreak omni}}
\newcommand{\omniSmall}{\texttt{jina-\allowbreak embeddings-\allowbreak v5-\allowbreak omni-\allowbreak small}}
\newcommand{\omniNano}{\texttt{jina-\allowbreak embeddings-\allowbreak v5-\allowbreak omni-\allowbreak nano}}
\newcommand{\omniSmallRetrieval}{\texttt{jina-\allowbreak embeddings-\allowbreak v5-\allowbreak omni-\allowbreak small-\allowbreak retrieval}}
\newcommand{\omniNanoRetrieval}{\texttt{jina-\allowbreak embeddings-\allowbreak v5-\allowbreak omni-\allowbreak nano-\allowbreak retrieval}}
\newcommand{\textSmall}{\texttt{jina-\allowbreak embeddings-\allowbreak v5-\allowbreak text-\allowbreak small}}
\newcommand{\textNano}{\texttt{jina-\allowbreak embeddings-\allowbreak v5-\allowbreak text-\allowbreak nano}}
\newcolumntype{L}[1]{>{\raggedright\arraybackslash}p{#1}}

\title{\omniBase{}: Geometry-preserving Embeddings via Locked Aligned Towers}

\author{Florian Hönicke}
\authornote{Principal contributor.}
\affiliation{\institution{Jina by Elastic}\city{}\country{}}
\email{research@jina.ai}

\author{Michael Günther}
\affiliation{\institution{Jina by Elastic}\city{}\country{}}
\email{research@jina.ai}

\author{Andreas Koukounas}
\affiliation{\institution{Jina by Elastic}\city{}\country{}}
\email{research@jina.ai}

\author{Mohammad Kalim Akram}
\affiliation{\institution{Jina by Elastic}\city{}\country{}}
\email{research@jina.ai}

\author{Saba Sturua}
\affiliation{\institution{Jina by Elastic}\city{}\country{}}
\email{research@jina.ai}

\author{Han Xiao}
\affiliation{\institution{Jina by Elastic}\city{}\country{}}
\email{research@jina.ai}

\renewcommand{\shortauthors}{Hönicke et al.}

\begin{document}

\begin{abstract}
In this work, we introduce GELATO (Geometry-preserving Embeddings via Locked Aligned TOwers), a novel approach to multimodal embedding models.
We build on the VLM-style architecture, in which non-text encoders are adapted to produce input for a language model, which in turn generates embeddings for all varieties of input. 
We present the result: the \omniBase{} suite, a pair of models that encode text, image, audio, and video input into a single semantic embedding space. 
GELATO extends the two Jina Embeddings v5 Text models to support additional modality by adding encoders for images and audio.
The backbone text embedding models and the added non-text modality encoders remain frozen. 
We only trained the connecting components, representing 0.35\% of the total weights of the joint model. 
Training is therefore much more efficient than full-parameter retraining. 
Additionally, the language model remains effectively unaltered: the text-encoder weights are bit-identical to the Jina Embeddings v5 Text models.
Our evaluations show that GELATO produces results competitive with substantially larger, state-of-the-art multimodal embedding models across text, image, and audio benchmarks.
\end{abstract}



\maketitle
\pagestyle{plain}

\section{Introduction}
\label{sec:introduction}
Text embedding models anchor retrieval, retrieval-augmented generation (RAG)~\cite{rag}, and classification pipelines whose vector indexes depend on a stable embedding geometry.
At the same time, search workloads increasingly require images, including screenshots, page scans, infographics, and other rendered media; audio, such as speech, music, and natural sounds; as well as video, to be queried alongside text.~\cite{mieb,vidore,mmeb,maeb}
%This creates a tension: multimodal embedders should broaden the modality coverage of retrieval systems, but changing the text embedding space invalidates existing text indexes and makes comparisons to the source text-only embedding model ambiguous.

\begin{figure}[t]
  \centering
  \vspace{0.75\baselineskip}
  \includegraphics[width=\columnwidth]{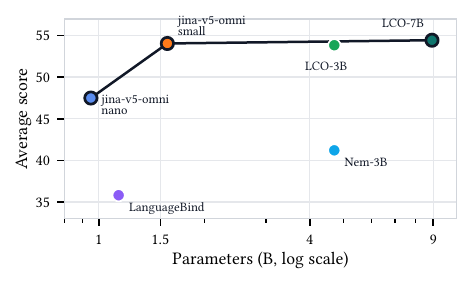}
  \caption{Average performance across multimodal embedding tasks versus model parameter count (see Table~\ref{tab:omni-frontier}).}
  \Description{A one-column frontier chart with Table 1 average scores for six open-weight omni models: jina-v5-omni-nano, jina-v5-omni-small, LanguageBind, LCO-3B, LCO-7B, and Nem-3B.}
  \label{fig:omni-average-frontier}
\end{figure}

\begin{figure*}[!t]
\centering
\includegraphics[width=\textwidth]{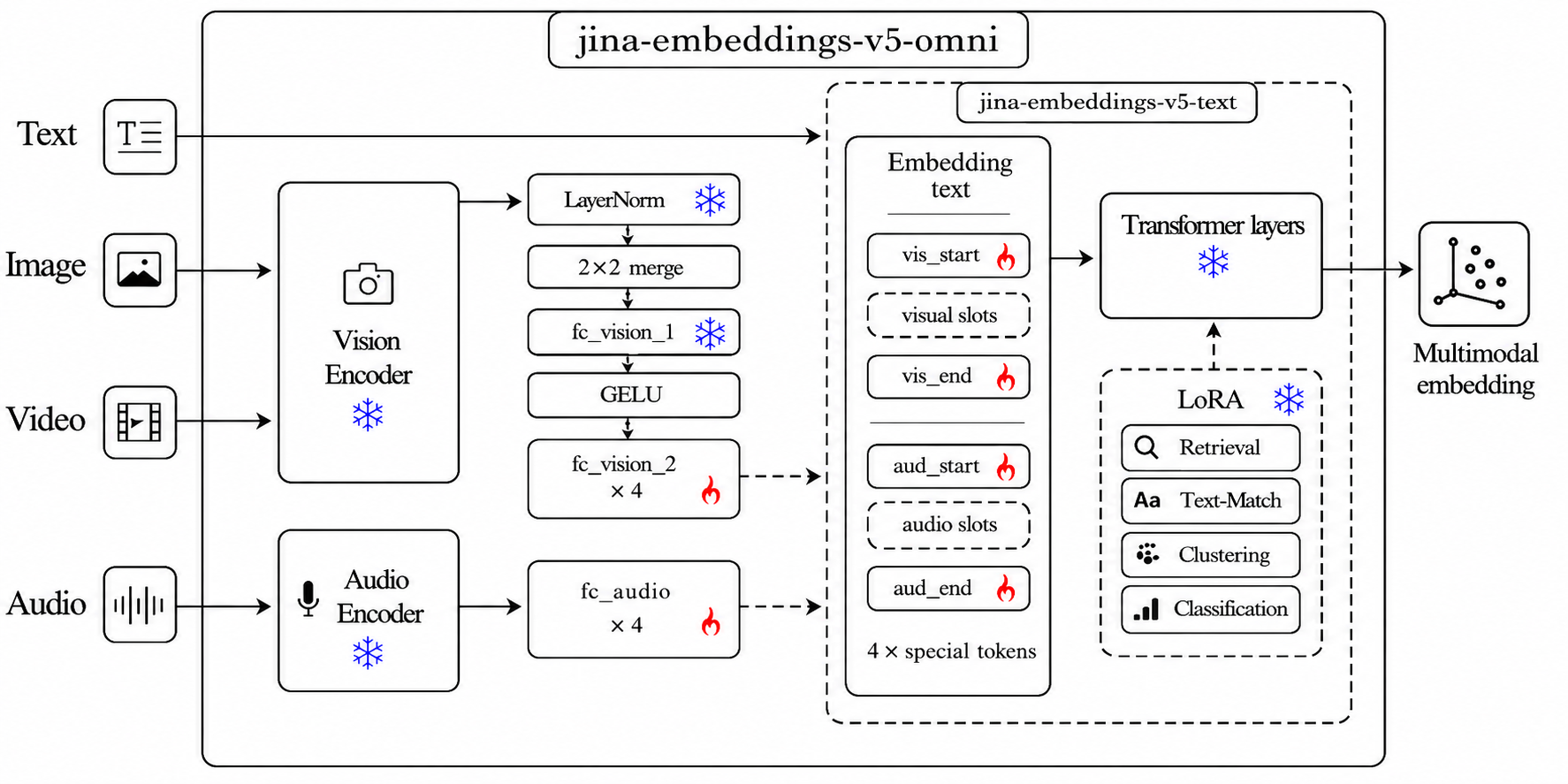}
\caption{Architecture of \omniBase{} (\omniSmall{} shown; \omniNano{} uses a smaller ViT and LLaVA-style tokens).
Frozen towers feed trainable modality projectors into the frozen text backbone; task-specific exports select one projector/delimiter set and the matching LoRA adapter.}
\label{fig:architecture}
\end{figure*}

We present \omniBase{}, a pair of models that extends a text embedding backbone to image, video, and audio while leaving the model entirely unchanged for text inputs. The two models differ substantially in size: \omniNano{} is based on \textNano{}, with 0.24B parameters in its base text-only model, and \omniSmall{}, based on \textSmall{} with 0.67B parameters.~\cite{jina-v5-text} 
The two base models have already been trained for high-performance text embeddings, using LoRA adapters to optimize them for multiple tasks: retrieval, text-matching, clustering, and classification. 

To add support for non-text modalities, we integrate:
\begin{itemize}
    \item Vision encoders from Qwen3.5-2B and Qwen3.5-0.8B~\cite{qwen3.5}, which have been adapted from SigLIP2 So400m and SigLIP2 Base respectively.~\cite{siglip2}
    \item The Qwen2.5-Omni audio encoder,~\cite{qwen2.5-omni} which has been adapted from Whisper-large-v3.~\cite{whisper}
\end{itemize}

The core idea of GELATO is to use independently pretrained, language-aligned encoders and align them to text embedding models through small trainable projectors rather than jointly retraining them. This makes it possible to readily construct modular multimodal embedding models while minimizing added parameters and additional training.

\paragraph{Contributions.}
\begin{enumerate}
  \item We describe GELATO and apply it in the construction of the \omniBase{} model suite by extending the Jina Embeddings v5 Text suite to support other media.
  \item We contribute to the open embedding ecosystem by releasing the \omniBase{} model collection\footnote{\href{https://huggingface.co/collections/jinaai/jina-embeddings-v5-omni-69f336b985c156b1d757029e}{Jina Embeddings v5 Omni Hugging Face collection}.}, comprising two base models and eight task-specific variants for retrieval, classification, clustering, and text-matching across Small and Nano scales.
  \item We evaluate \omniBase{} and comparable models across a range of standard benchmarks, and show that GELATO produces competitive results. (See Figure~\ref{fig:omni-average-frontier}.)
  \item We analyze the design rules behind GELATO through ablations on projector training, encoder choice, and Matryoshka truncation, and separately quantify training efficiency.
\end{enumerate}

\section{Related Work}
\label{sec:related}
Text-only embedding models are long established for retrieval and RAG systems, from bidirectional encoders such as Sentence-BERT~\cite{sentence-bert} and GTE-Qwen2~\cite{gte-qwen2} to LLM-based text-only embedding models such as E5-Mistral~\cite{e5-mistral} and NV-Embed~\cite{nv-embed}.
Jina Embeddings v5 Text~\cite{jina-v5-text} draws on this tradition: a state-of-the-art model family with task-conditioned LoRA adapters and support for truncation with low performance loss due to Matryoshka representation learning~\cite{matryoshka}.

CLIP~\cite{clip} established contrastive image--text embedding with separately encoded image and text towers, and SigLIP~\cite{siglip}, SigLIP2~\cite{siglip2}, and EVA-CLIP~\cite{eva-clip} refine this paradigm through improved losses, data, and visual training recipes.
ImageBind~\cite{imagebind} extends contrastive alignment to additional modalities.
Jina CLIP v1/v2~\cite{jinaclip-v1,jinaclip-v2} maintains text-embedding performance in CLIP-style models, while supporting other media.
However, contrastively-trained multimodal embedders suffer from a gap between modality-specific regions of the shared representation space~\cite{modality-gap}.

VLM-style architectures tackle this challenge by passing the outputs of non-text media encoders through the same language model as the text token representations. 
These models, including LLaVA~\cite{llava}, BLIP-2~\cite{blip2}, Qwen2-VL~\cite{qwen2-vl}, and Qwen3-VL~\cite{qwen3-vl}, use projectors or connector modules to connect the encoders to the language model.
Embedding models derived from VLMs, like E5-V~\cite{e5-v}, GME~\cite{gme-qwen2-vl}, and Qwen3-VL-Embedding~\cite{qwen3-vl-embedding}, demonstrate strong multimodal retrieval performance, but involve adapting the language model, non-text media encoders, or both.

Omni-style systems train or align multiple modalities jointly, supporting video and audio in addition to images, for example, E5-Omni~\cite{e5-omni}, WAVE~\cite{wave}, and LCO-Embedding-Omni~\cite{lco-embedding}. 

We take note of previous work in frozen-tower methods based on the CLIP architecture, such as LiT~\cite{lit} and Nomic Embed Vision~\cite{nomic-embed-vision}, which freeze the text encoder while adapting the other media towers.
The most relevant prior work is the visual-module-plugin family.
MARVEL~\cite{marvel} adds a CLIP visual encoder with a linear projection layer to the frozen text retriever T5-ANCE, then finetunes the text retriever end-to-end for multi-modal retrieval.
VISTA~\cite{vista} extends a frozen text embedding model with a trainable ViT image tokenizer, keeping the text encoder frozen throughout.
To the best of our knowledge, no previously published work extends a frozen text embedding model to image, video, and audio jointly while keeping all media encoders frozen and training only a single linear projector layer per modality together with a small set of modality delimiter token embeddings.

\section{Architecture}
\label{sec:method}
Figure~\ref{fig:architecture} summarizes the architecture of the \omniBase{} models.
We extend Jina Embeddings v5 Text from text-only embedding to vision and audio by adding scale-matched Qwen3.5 vision encoders\footnote{\omniSmall{} uses \texttt{Qwen/Qwen3.5-2B}; \omniNano{} uses \texttt{Qwen/Qwen3.5-0.8B}.} and the Qwen2.5-Omni audio encoder to the same text-sequence backbone.
We chose encoders from trained multimodal language systems rather than bare perceptual encoders such as SigLIP2 or Whisper-large because prior work shows that visual and audio features need explicit language-space alignment or natural-language supervision before they transfer reliably to text-conditioned multimodal tasks~\cite{comp,clap,qwen3.5,qwen2.5-omni}.
The text processing path of \omniBase{} shares the frozen text-encoder weights of Jina Embeddings v5 Text bit-for-bit: token embeddings pass through the frozen text transformer, the inherited task LoRA adapter is applied, and the final embedding is produced by last-token pooling and L2 normalization.

\subsection{Projectors}
\label{sec:method:projectors}

\omniBase{} uses image and audio encoders extracted from Qwen3.5 and Qwen2.5-Omni, respectively. Because their output dimensions do not match Jina Embeddings v5 Text's input, we replace the source projection layers with new projectors that map into the text hidden space.
For audio, we inserted a randomly-initialized \texttt{fc\_audio} layer that projects the encoder's native $1280$ dimension output into \omniSmall{}'s $1024$-dimension input space and \omniNano{}'s $768$-dimension one.

We write each fully connected layer as the same affine map
\begin{equation*}
\ell_{W,\mathbf{b}}(\mathbf{x}) = W\mathbf{x} + \mathbf{b},
\label{eq:linear_layer}
\end{equation*}
with layer-specific weights and bias.
Thus \texttt{fc\_vision\_1} is $\ell_{W_{\text{v1}},\mathbf{b}_{\text{v1}}}$, \texttt{fc\_vision\_2} is $\ell_{W_{\text{v2}},\mathbf{b}_{\text{v2}}}$, and \texttt{fc\_audio} is $\ell_{W_{\text{aud}},\mathbf{b}_{\text{aud}}}$.

For vision, the Qwen3.5 visual projector converts ViT patch tokens into text-token features by applying LayerNorm, a $2{\times}2$ spatial merge, \texttt{fc\_vision\_1}, GELU, and \texttt{fc\_vision\_2}.
Here, LayerNorm denotes feature normalization on the ViT patch tokens.
The $2{\times}2$ spatial merge is a fixed space-to-depth (pixel-unshuffle) rearrangement that concatenates four neighboring patch embeddings into one $4d_{\text{vit}}$ vector, reducing the spatial token count by $4\times$; it is the inverse direction of pixel shuffle/sub-pixel rearrangement~\cite{pixel-shuffle} and follows Qwen's visual-merger design~\cite{qwen2-vl,qwen3.5}.
For each group of four neighboring patch tokens $\mathbf{V}_i = [\mathbf{v}_{i,1}, \ldots, \mathbf{v}_{i,4}] \in \mathbb{R}^{4 \times d_{\text{vit}}}$, the vision projector produces
\begin{equation*}
\begin{aligned}
\mathbf{m}^{(i)}_{\text{vis}} &=
\bigl[
\text{LayerNorm}(\mathbf{v}_{i,1});
\ldots;
\text{LayerNorm}(\mathbf{v}_{i,4})
\bigr] \in \mathbb{R}^{4d_{\text{vit}}},\\
\mathbf{z}^{(i)}_{\text{vis}} &=
\text{GELU}\!\left(\ell_{W_{\text{v1}},\mathbf{b}_{\text{v1}}}(\mathbf{m}^{(i)}_{\text{vis}})\right),\\
\mathbf{h}^{(i)}_{\text{vis}} &=
\ell_{W_{\text{v2}},\mathbf{b}_{\text{v2}}}(\mathbf{z}^{(i)}_{\text{vis}}),
\qquad i = 1,\ldots,N_{\text{vis}}.
\end{aligned}
\label{eq:vis_proj}
\end{equation*}

Only \texttt{fc\_vision\_2} performs the dimension-specific projection into a text hidden space: in the 2B source checkpoint it maps $4096{\to}2048$ into the Qwen3.5-2B text hidden dimension, and in the 0.8B source checkpoint it maps $3072{\to}1024$ into the Qwen3.5-0.8B text hidden dimension.
These targets do not match Small's $1024$-dimensional or Nano's $768$-dimensional Jina text backbone, so we keep LayerNorm and \texttt{fc\_vision\_1} frozen but replace \texttt{fc\_vision\_2} with a randomly initialized $4096{\to}1024$ layer for Small and $3072{\to}768$ layer for Nano.

Let $\mathbf{A} = [\mathbf{a}_1, \ldots, \mathbf{a}_K] \in \mathbb{R}^{K \times 1280}$ denote the frozen Qwen2.5-Omni audio encoder states for an input with $K$ audio tokens.
Each audio token is independently projected into the Jina text hidden dimension by \texttt{fc\_audio}
\begin{equation*}
\mathbf{h}^{(i)}_{\text{aud}} =
\ell_{W_{\text{aud}},\mathbf{b}_{\text{aud}}}(\mathbf{a}_i),
\qquad i = 1,\ldots,K,
\label{eq:aud_proj}
\end{equation*}
where $W_{\text{aud}} \in \mathbb{R}^{d_{\text{text}} \times 1280}$ and $d_{\text{text}} \in \{1024,768\}$ for Small and Nano.

\subsection{Input Sequence Construction}
\label{sec:method:input-sequence}

Each input is serialized as one sequence of tokens.
Text remains ordinary text tokens; non-text modalities are represented by placeholder runs inside modality delimiters.
An image is encoded as
\begin{equation*}
\texttt{<|vision\_start|>} \;\; \underbrace{\texttt{<|image\_pad|>} \times N}_{\text{visual slots}} \;\; \texttt{<|vision\_end|>}
\label{eq:img_seq}
\end{equation*}
with $N$ visual slots.
An audio input is encoded as
\begin{equation*}
\texttt{<|audio\_start|>} \;\; \underbrace{\texttt{<|audio\_pad|>} \times K}_{\text{audio slots}} \;\; \texttt{<|audio\_end|>}
\label{eq:aud_seq}
\end{equation*}
with $K$ audio slots.
A video is a concatenation of one visual segment per sampled frame:
\begin{equation*}
\big\Vert_{f=1}^{F}
\left(
\texttt{<|vision\_start|>} \;\; \underbrace{\texttt{<|video\_pad|>} \times S_f}_{\text{frame } f \text{ slots}} \;\; \texttt{<|vision\_end|>}
\right),
\label{eq:video_seq}
\end{equation*}
where $\Vert$ denotes sequence concatenation.
If a video contains an audio track, the extracted audio segment precedes the frame sequence:
\begin{equation*}
\mathbf{s}_{\text{aud}} \Vert \mathbf{s}_{\text{vid}}.
\label{eq:video_audio_seq}
\end{equation*}
Here, $\mathbf{s}_{\text{aud}}$ is the audio sequence above and $\mathbf{s}_{\text{vid}}$ is the video-frame sequence.
For mixed-modality inputs, text spans and modality segments are concatenated in document order.

\subsection{Trainable Parameters}
\label{sec:method:trainable-parameters}

The trainable set is \texttt{fc\_vision\_2}, \texttt{fc\_audio}, and the modality-delimiter embeddings.
\omniSmall{} learns the vision and audio start/end delimiter embeddings used in Section~\ref{sec:method:input-sequence}; \omniNano{} learns only the audio start/end delimiter embeddings.
The image, video, and audio placeholder positions are overwritten by projected encoder features rather than learned as standalone token embeddings.
Projector and delimiter-token training is run separately for retrieval, text-matching, clustering, and classification, while the text transformer, encoder towers, LayerNorm/\texttt{fc\_vision\_1} vision-projector weights, and inherited LoRA adapters stay frozen.
The base package stores four such task-specific sets alongside the inherited LoRA adapters.

\subsection{Dynamic Weight Loading}
\label{sec:method:dynamic-loading}

Jina Embeddings v5 Text already uses dynamic adapter selection to route retrieval, classification, clustering, and text-matching inputs through the corresponding task adapter.
We extend the same task-selection mechanism to the multimodal weights: the selected task variant determines which LoRA adapter, \texttt{fc\_vision\_2}, \texttt{fc\_audio}, and learned special text-token embeddings are loaded or activated.
The task-specific projector and delimiter-token weights therefore follow the same task-specific variation as Jina Embeddings v5 Text.
Separately, the model exposes a modality attribute that controls which frozen modality towers are instantiated: text-only loading omits both vision and audio towers, vision-only loading omits the audio tower and \texttt{fc\_audio}, audio-only loading omits the vision tower and vision projector, and omni loading keeps both vision and audio towers.

\section{Training}
\label{sec:training}
\label{sec:training:loss}

Projector training uses bidirectional in-batch InfoNCE with Matryoshka representation learning.
For a batch of $B$ paired examples $\{(\ell_i,r_i)\}_{i=1}^{B}$, let $\mathbf{u}_i$ and $\mathbf{v}_i$ be the left and right embeddings, and let $\mathbf{u}_{i,1:k}$ denote the first $k$ dimensions.
With temperature $\tau=0.02$,
\begin{equation*}
\begin{aligned}
s_{ij}^{(k)}&=\frac{\cos(\mathbf{u}_{i,1:k},\mathbf{v}_{j,1:k})}{\tau},\\
p_{\ell\to r}^{(k)}(j|i)&=\frac{\exp(s_{ij}^{(k)})}{\sum_{m=1}^{B}\exp(s_{im}^{(k)})},\\
p_{r\to \ell}^{(k)}(j|i)&=\frac{\exp(s_{ji}^{(k)})}{\sum_{m=1}^{B}\exp(s_{mi}^{(k)})}.
\end{aligned}
\end{equation*}
\begin{equation*}
\mathcal{L}_{\mathrm{NCE}}^{(k)}=-\frac{1}{2B}\sum_{i=1}^{B}\left[\log p_{\ell\to r}^{(k)}(i|i)+\log p_{r\to \ell}^{(k)}(i|i)\right].
\label{eq:infonce}
\end{equation*}
The training loss sums this term over Matryoshka prefix dimensions,
\begin{equation*}
\begin{aligned}
\mathcal{L}&=\sum_{k\in\mathcal{K}}\mathcal{L}_{\mathrm{NCE}}^{(k)},\\
\mathcal{K}_{\mathrm{Small}}&=\{32,64,128,256,512,768,1024\},\\
\mathcal{K}_{\mathrm{Nano}}&=\{32,64,128,256,512,768\}.
\end{aligned}
\label{eq:matryoshka_loss}
\end{equation*}

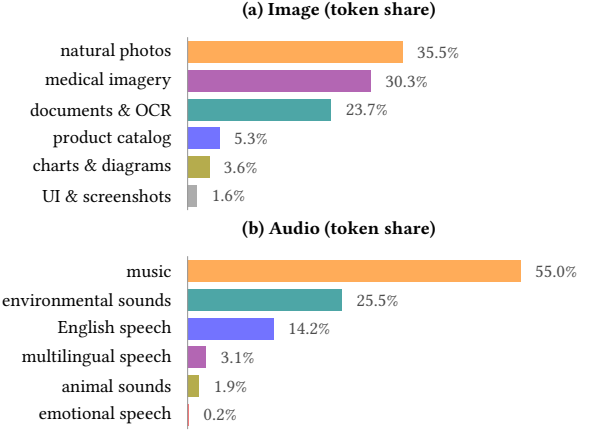
\begin{figure}[H]
\centering
% Stacked bar charts of averaged sampling rates across the four task variants.
% Each row: \bar{label}{percent}{fillcolor}
\begin{tikzpicture}[
    barlabel/.style={font=\footnotesize, anchor=east},
    barvalue/.style={font=\footnotesize, anchor=west, text=black!70},
    axisrule/.style={black!40, line width=0.4pt},
]
    \def\rowh{0.38}
    \def\scale{0.08} % 1% = 0.08 cm, so 50% = 4 cm
    \node[font=\footnotesize\bfseries] at (2.0, 0.55) {(a) Image (token share)};
    \foreach \lbl/\pct/\col [count=\i from 0] in {%
        natural photos/35.5/orange!65,%
        medical imagery/30.3/violet!60,%
        documents \& OCR/23.7/teal!70,%
        product catalog/5.3/blue!55,%
        charts \& diagrams/3.6/olive!70,%
        UI \& screenshots/1.6/gray!65%
    }{
        \pgfmathsetmacro{\y}{-\i*\rowh}
        \pgfmathsetmacro{\w}{\pct*\scale}
        \node[barlabel] at (-0.1, \y) {\lbl};
        \fill[\col] (0, \y-0.14) rectangle (\w, \y+0.14);
        \node[barvalue] at (\w+0.08, \y) {\pct\%};
    }
    \draw[axisrule] (0, 0.2) -- (0, -5.5*\rowh);

    \begin{scope}[yshift=-2.9cm]
    \node[font=\footnotesize\bfseries] at (2.0, 0.55) {(b) Audio (token share)};
    \foreach \lbl/\pct/\col [count=\i from 0] in {%
        music/55.0/orange!65,%
        environmental sounds/25.5/teal!70,%
        English speech/14.2/blue!55,%
        multilingual speech/3.1/violet!60,%
        animal sounds/1.9/olive!70,%
        emotional speech/0.2/red!55%
    }{
        \pgfmathsetmacro{\y}{-\i*\rowh}
        \pgfmathsetmacro{\w}{\pct*\scale}
        \node[barlabel] at (-0.1, \y) {\lbl};
        \fill[\col] (0, \y-0.14) rectangle (\w, \y+0.14);
        \node[barvalue] at (\w+0.08, \y) {\pct\%};
    }
    \draw[axisrule] (0, 0.2) -- (0, -5.5*\rowh);
    \end{scope}

\end{tikzpicture}
\caption{Distribution of input \emph{tokens} across semantic data types, averaged over the four task-specific checkpoints.}
\label{fig:data-distribution}
\end{figure}

We use the AdamW optimizer~\cite{adamw} with $\beta_1{=}0.9$, $\beta_2{=}0.999$, weight decay $0.01$, and global gradient clipping at $\lVert \nabla \rVert_2 \le 1$.
The learning rate is $2{\cdot}10^{-4}$ with $500$ linear warmup steps.
Training uses bf16 mixed precision and distributed data parallelism across $4$ NVIDIA H100 GPUs, with global batch size $256$ paired examples.
Projector training is run separately per modality and per task: the vision projector (\texttt{fc\_vision\_2} plus the vision modality-delimiter embeddings, where applicable) and the audio projector (\texttt{fc\_audio} plus the audio modality-delimiter embeddings) are trained in independent runs, each one using the corresponding frozen LoRA adapter inherited from Jina Embeddings v5 Text and a task-matched source mixture.
Across both model sizes, the four task variants (retrieval, classification, clustering, text-matching), and the two modalities (vision, audio), this yields $2 \times 4 \times 2 = 16$ projector-training runs in total.
Each run is trained for $15\,000$ optimizer steps.
Each batch contains examples from one source dataset sampled by mixture weight.
Figure~\ref{fig:data-distribution} summarizes the shared projector-training mixture by token share across semantic data types.
The mixture is full of text-rich and complex images like scans and diagrams, matching practical enterprise search and RAG systems that operate over real-world multimodal documents whose layout, images, and OCR/parsing stages affect retrieval quality~\cite{rag,visrag}.

\section{Evaluation}
\label{sec:experiments}
\label{sec:experiments:benchmarks}

We describe each evaluation suite by the types of tasks it covers:
\begin{itemize}
    \item \textbf{Images:} The Massive Image Embedding Benchmark (MIEB)~\cite{mieb} covers classification, clustering, visual semantic textual similarity (STS), retrieval, document retrieval, compositional reasoning, and vision-centric tasks.
    \item \textbf{Video:} The Massive Multimodal Embedding Benchmark (MMEB)~\cite{mmeb} provides a video evaluation suite, MMEB-Video, covering classification, VQA, retrieval, and moment-retrieval sub-tasks.
    \item \textbf{Audio:} The Massive Audio Embedding Benchmark (MAEB)~\cite{maeb} covers audio--text and audio-centric embedding quality, grouped by task type (retrieval, classification, clustering, text-matching).
    \item \textbf{Text:} The Massive Multilingual Text Embedding Benchmark (MMTEB)~\cite{mteb} evaluates text-only embedding quality across retrieval, classification, clustering, semantic textual similarity, reranking, and pair-classification tasks.
\textbf{Documents.} We report ViDoRe~\cite{vidore} page-level retrieval, where embeddings must capture fine layout and small text.
\end{itemize}

For text, we report the published MMTEB scores for the inherited Jina Embeddings v5 Text encoders, which \omniBase{} shares bit-for-bit.~\cite{jina-v5-text}

Our baselines for comparison consist of open-weight omni-style models with support for the same media types: LanguageBind, Omni-Embed-Nemotron-3B, LCO-Embedding-Omni-3B, and LCO-Embedding-Omni-7B. 
It also includes some task-matched specialized models: CLIP/SigLIP-style and VLM-derived embedders for vision, Whisper/CLAP-style embedders for audio, and VLM/video embedding models for video.
Parameter counts are task-path specific: summaries for omni-style models count all compared modalities, while modality-specific rows count only the encoders needed for that task.

\begin{table*}[th]
  \caption{Open-weight omni-style model scores on selected evaluation subsets.
  Text uses MMTEB; Image, Video, and Audio use aggregate MIEB, MMEB-Video subset\textsuperscript{a}, and MAEB scores, respectively.}
  \label{tab:omni-frontier}
  \centering
  \small
  \setlength{\tabcolsep}{5pt}
  \begin{tabular*}{\textwidth}{@{\extracolsep{\fill}}lcccccc@{}}
    \toprule
    Model & Params (B) & Text & Image & Video & Audio & Avg \\
    \midrule
    \omniNano{} & 0.95 & 65.52 & 47.87 & 26.87 & 49.69 & 47.49 \\
    \texttt{LanguageBind} & 1.14 & 27.34 & 47.80 & \textbf{48.06} & 20.08 & 35.82 \\
    \omniSmall{} & 1.57 & \textbf{67.00} & 58.00 & 41.20 & 49.96 & 54.04 \\
    \texttt{Omni-Embed-Nemotron-3B} & 4.70 & 47.64 & 44.47 & 24.46 & 48.27 & 41.21 \\
    \texttt{LCO-Embedding-Omni-3B} & 4.70 & 57.55 & 58.42 & 46.84 & \textbf{52.51} & 53.83 \\
    \texttt{LCO-Embedding-Omni-7B} & 8.93 & 59.31 & \textbf{58.64} & 47.41 & 52.37 & \textbf{54.43} \\
    \bottomrule
  \end{tabular*}
  \\[0.4em]
  \begin{minipage}{0.97\textwidth}
  \footnotesize
  \textsuperscript{a} MMEB-Video subset: Breakfast, MSR-VTT, EgoSchema, HMDB51, UCF101, MSVD, SmthSmthV2, DiDeMo, and K700.
  Params count the loaded parameters needed for text, image, video, and audio requests; LanguageBind counts one shared language encoder plus the Image, Video\_FT, and Audio\_FT modality paths, not duplicate text copies shipped across the separate checkpoints.
  Avg averages the displayed numeric columns.
  \end{minipage}
\end{table*}

\begin{table}[th]
  \caption{Document-retrieval scores on the ViDoRe-in-MIEB subset.}
  \label{tab:omni-doc-retrieval}
  \centering
  \small
  \setlength{\tabcolsep}{3pt}
  \begin{tabular}{@{}lcc@{}}
    \toprule
    Model & Params$^{*}$ (B) & Document retrieval \\
    \midrule
    \omniNano{} & 0.31 & 79.25 \\
    \texttt{LanguageBind} & 0.43 & 37.33 \\
    \omniSmall{} & 0.92 & 79.25 \\
    \texttt{LCO-Embedding-Omni-3B} & 4.07 & 78.24 \\
    \texttt{Omni-Embed-Nemotron-3B} & 4.70 & \textbf{85.64} \\
    \texttt{LCO-Embedding-Omni-7B} & 8.93 & 80.32 \\
    \bottomrule
  \end{tabular}
  \\[0.4em]
  \begin{minipage}{0.97\columnwidth}
  \footnotesize
  \textsuperscript{*}Text+image path parameters for document retrieval; audio/video encoders are not counted.
  Subset tasks: DocVQA, InfoVQA, TabFQuAD, TAT-DQA, ArxivQA, ShiftProject, SyntheticDocQA-AI, SyntheticDocQA-Energy, SyntheticDocQA-HealthcareIndustry, and SyntheticDocQA-GovernmentReports.
  \end{minipage}
\end{table}

\subsection{Results}
\label{sec:experiments:summary}

Table~\ref{tab:omni-frontier} shows that \omniSmall{} has the strongest text-only performance and the best overall score among models below $5$B parameters.
Its $54.04$ four-modality average is slightly above LCO-Embedding-Omni-3B ($53.83$) and below only the larger LCO-Embedding-Omni-7B score of $54.43$, among comparable omni-style models.
The same table also contains comparisons by modality. \omniSmall{} is very strong on text and competitive on images and audio, but video performance lags significantly compared to the baseline models.

Table~\ref{tab:omni-doc-retrieval} shows that both \omniNano{} and \omniSmall{} have strong visual document retrieval performance.
\omniSmall{} scores $79.25$ with $0.92$B active text+image-path parameters, above LCO-Embedding-Omni-3B ($78.24$) and close to LCO-Embedding-Omni-7B ($80.32$).
\omniNano{} matches that with $79.25$ at $0.31$B active parameters, also surpassing LCO-Embedding-Omni-3B ($78.24$) at less than a tenth the parameter count.

Table~\ref{tab:benchmark-summary} gives a detailed breakdown across multiple benchmarks. The strongest \omniSmall{} performances are for image classification, image clustering, visual STS, multilingual image retrieval, and audio classification, while generic image retrieval, MMEB-Video, and audio clustering remain weaker.

\begin{figure}[t]
  \centering
  \includegraphics[width=\columnwidth]{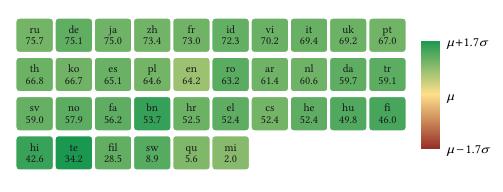}
  \caption{XM3600 image-language comparison.
  Tiles show \texttt{jina-v5-omni-small}; color is deviation from a five-model language mean.}
  \Description{XM3600 language tiles for jina-v5-omni-small compared with Nano, LanguageBind, LCO-3B, and Nem-3B; LCO-7B is discussed by aggregate score.}
  \label{fig:multilingual-comparison}
\end{figure}

\begin{figure}[t]
  \centering
  \includegraphics[width=\columnwidth]{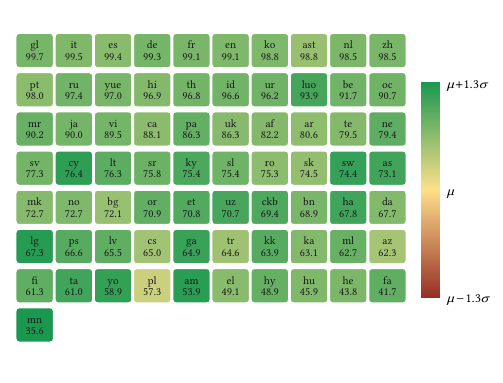}
  \caption{Per-language audio retrieval.
  Tiles show \texttt{jina-v5-omni-small} on shared CommonVoiceMini21/FLEURS languages; color is deviation from the mean of the baseline models.}
  \Description{Audio language tiles for jina-v5-omni-small compared with Nano, LCO-3B, Nem-3B, and LanguageBind Audio across the shared CommonVoiceMini21/FLEURS languages.}
  \label{fig:audio-language-comparison}
\end{figure}

Figures~\ref{fig:multilingual-comparison} and~\ref{fig:audio-language-comparison} show relative performance per language, compared to the average of the baseline models. Color indicates deviation from the five-model per-language mean for image-language and audio retrieval, respectively.
Figure~\ref{fig:multilingual-comparison} highlights the relatively strong performance of \omniSmall{} on languages other than English, while Figure~\ref{fig:audio-language-comparison} does the same for audio performance.

\subsection{Modality Geometry}
\label{sec:exp:modality-geometry}

%The frozen vision and audio encoders never observe the contrastive text-embedding loss, so the joint space is not guaranteed to align modalities to the same degree as a fully co-trained model.
Freezing the vision and audio encoders means that they are not contrastively trained the way a fully co-trained model would be, and therefore may not align the different modalities to the same degree.
We evaluate the resulting embedding geometry and retrieval quality on the MS-COCO Karpathy split ($2{,}000$ images $\times$ $5$ captions = $10{,}000$ caption candidates) and the Clotho v2 test split ($1{,}045$ audio--caption pairs, $1$ caption each), comparing the \omniBase{} models to the three co-trained omni baselines from Table~\ref{tab:benchmark-summary}: \texttt{LCO-Embedding-\allowbreak Omni-3B} ($3.7$B), \texttt{LCO-Embedding-Omni-7B} ($8.93$B), and \texttt{Omni-Embed-\allowbreak Nemotron-3B} ($4.7$B).
We exclude the fourth omni-style baseline of Table~\ref{tab:benchmark-summary}, \texttt{LanguageBind}, because its per-modality CLIP-style architecture is structurally different from the unified-decoder pattern used by our omni models, so its modality-gap and retrieval-margin readings are not comparable.

\begin{figure*}[t]
  \centering
  \includegraphics[width=\textwidth]{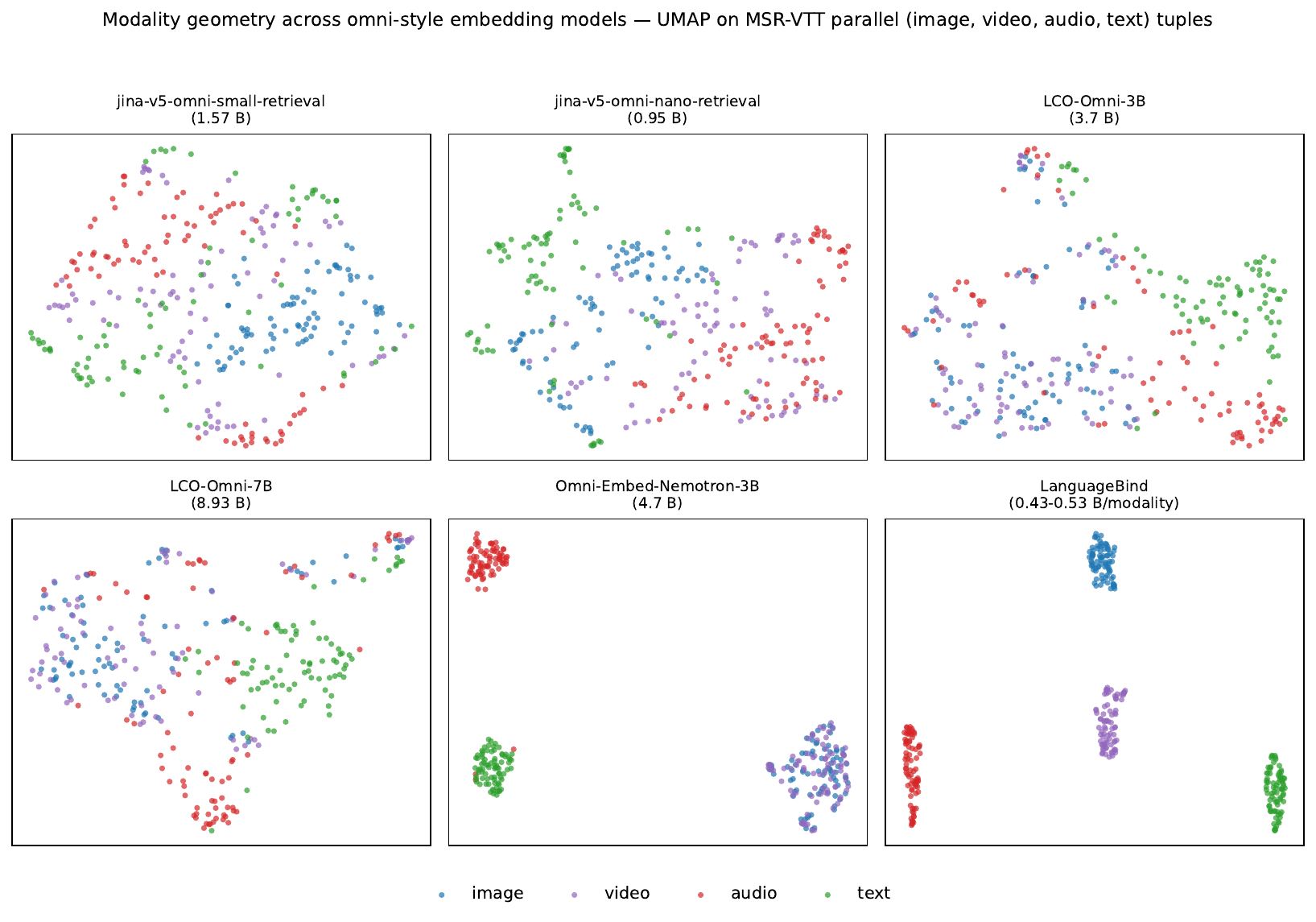}
  \caption{2D UMAP of embeddings on $80$ parallel MSR-VTT clips per model.
  Each row of source data contributes four embeddings (image: middle frame; video: full clip; audio: extracted track; text: caption); colours mark modalities.
  UMAP is fit per model on its own $320$ ($80\times4$) points (\texttt{cosine} metric, \texttt{n\_neighbors}=$15$, \texttt{min\_dist}=$0.1$, \texttt{random\_state}=$42$).
  \texttt{LanguageBind}'s frozen per-modality towers produce the canonical modality-gap pattern; the unified-decoder omni models, including the frozen-tower \omniSmall{} and \omniNano{}, produce interleaved geometry.}
  \Description{Six-panel UMAP grid comparing modality clustering across omni embedding models.}
  \label{fig:modality-umap}
\end{figure*}

All embeddings are $L_2$-normalised, and each model encodes texts and media in its natural retrieval protocol.
Following~\cite{modality-gap}, we report \emph{centroid distance} $\lVert\mu_A - \mu_B\rVert_2$ between modality means as the direct modality-gap measure, complemented by \emph{alignment lift} (paired minus random-pair mean cosine) and the UMAP geometry of Figure~\ref{fig:modality-umap}.
We also report cross-modal \emph{Recall@1} in both directions on the multi-caption MS-COCO protocol (image$\rightarrow$text and text$\rightarrow$image; for audio, audio$\rightarrow$text and text$\rightarrow$audio) as a coupling signal --- these directions probe whether matched pairs are retrievable across modalities, not the modality gap itself, which would require a mixed-modality target pool (e.g. text$\rightarrow$\{text, image\}).
Between-model R@$1$ differences carry confidence intervals from a paired bootstrap ($2{,}000$ resamples) over per-query correctness; these CIs are tabulated alongside the headline R@$1$ values in Table~\ref{tab:modality-geometry}.
All measurements use the retrieval task-specific variants; classification, clustering, and text-matching variants carry different LoRA adapters and shift the absolute cosines.

\begin{table*}[t]
\centering
\small
\caption{Modality geometry and retrieval across omni embedding models on MS-COCO (image--text, I-T, $2{,}000$ images $\times$ $5$ captions) and Clotho v2 (audio--text, A-T, $1{,}045$ audio $\times$ $1$ caption). $\rightarrow$ denotes primary $\rightarrow$ text retrieval (image$\rightarrow$text / audio$\rightarrow$text); $\leftarrow$ denotes text $\rightarrow$ primary. All cosines are between $L_2$-normalised embeddings. Lift is paired minus random-pair mean cosine. R@$1$ in percent. Rows are ordered by mean R@$1$ across directions within each pair. Bold marks the column winner per pair.}
\label{tab:modality-geometry}
\setlength{\tabcolsep}{3pt}
\begin{tabular}{@{}llcccccc@{}}
\toprule
Model & Pair & Params & Cent.\ $L_2$ & Paired & Lift & R@$1\,\rightarrow$ & R@$1\,\leftarrow$ \\
\midrule
\texttt{LCO-Omni-7B}                 & I-T & $8.93$B & 0.46 & 0.55 & \textbf{0.42} & \textbf{74.0} & \textbf{63.6} \\
\texttt{LCO-Omni-3B}                 & I-T & $3.7$B  & 0.43 & 0.57 & 0.41 & 71.6 & 58.0 \\
\omniSmallRetrieval{}                & I-T & $1.57$B & 0.71 & 0.40 & 0.30 & 68.0 & 57.0 \\
\omniNanoRetrieval{}                 & I-T & $0.95$B & 0.54 & 0.35 & 0.25 & 36.6 & 27.7 \\
\texttt{Omni-Embed-Nemotron-3B}      & I-T & $4.7$B  & 0.92 & 0.35 & 0.14 & 23.1 & \phantom{0}1.4 \\
\midrule
\texttt{LCO-Omni-7B}                 & A-T & $8.93$B & 0.39 & 0.56 & \textbf{0.28} & \textbf{27.5} & \textbf{29.8} \\
\texttt{LCO-Omni-3B}                 & A-T & $3.7$B  & 0.39 & 0.56 & 0.27 & 24.5 & 28.5 \\
\omniSmallRetrieval{}                & A-T & $1.57$B & 0.64 & 0.32 & 0.17 & 16.3 & 15.2 \\
\omniNanoRetrieval{}                 & A-T & $0.95$B & 0.63 & 0.29 & 0.14 & 14.1 & 14.8 \\
\texttt{Omni-Embed-Nemotron-3B}      & A-T & $4.7$B & 0.56 & 0.53 & 0.11 & 10.1 & \phantom{0}9.6 \\
\bottomrule
\end{tabular}
\\[0.4em]
\begin{minipage}{0.97\textwidth}
\footnotesize
Paired-bootstrap ($2{,}000$ resamples) $95\%$ CIs for R@$1$ differences vs.\ \omniSmallRetrieval{}.
Image--text: vs.\ \texttt{LCO-Omni-3B} image$\rightarrow$text $[-5.8, -1.3]$, text$\rightarrow$image $[-2.0, +0.0]$; vs.\ \texttt{LCO-Omni-7B} image$\rightarrow$text $[-8.2, -3.9]$, text$\rightarrow$image $[-7.6, -5.7]$.
Audio--text: vs.\ \texttt{LCO-Omni-7B} audio$\rightarrow$text $[-14.3, -8.2]$, text$\rightarrow$audio $[-17.4, -11.7]$.
\end{minipage}
\end{table*}

\paragraph{Image--text.}
The LCO-Embedding family leads on MS-COCO Karpathy: \texttt{LCO-Omni-7B} ($8.93$B) at image$\rightarrow$text R@$1 = 74.0\%$ / text$\rightarrow$image $63.6\%$, \texttt{LCO-Omni-3B} ($3.7$B) at $71.6\%$ / $58.0\%$ (Table~\ref{tab:modality-geometry}).
Our \omniSmall{} variant ($1.57$B, frozen-tower) sits in third place at $68.0\%$ / $57.0\%$, statistically indistinguishable from \texttt{LCO-Omni-3B} on text$\rightarrow$image at the edge of significance, trailing it by $3.6\%$ on image$\rightarrow$text and trailing \texttt{LCO-Omni-7B} by $6$--$7\%$ in both directions; paired-bootstrap $95\%$ confidence intervals are tabulated in the Table~\ref{tab:modality-geometry} footnote.
At $18\%$ the parameter count of \texttt{LCO-Omni-7B} and $42\%$ of \texttt{LCO-Omni-3B}, the frozen-tower design closes most of the omni gap in both directions.
The \omniNano{} variant ($0.95$B) trails the leaders at $36.6\%$ / $27.7\%$, and \texttt{Omni-Embed-Nemo\-tron-3B} ($4.7$B, co-trained) collapses to $23.1\%$ / $1.4\%$. The low score for \texttt{Omni-Embed-Nemotron-3B} is evidence that parameter count alone does not determine outcome on this axis.

\paragraph{Audio--text.}
The LCO-Embedding family leads for this modality as well: \texttt{LCO-Omni-7B} at $27.5\%$ / $29.8\%$, \texttt{LCO-Omni-3B} at $24.5\%$ / $28.5\%$ (Table~\ref{tab:modality-geometry}).
Our \omniSmall{} variant reaches $16.3\%$ / $15.2\%$, trailing \texttt{LCO-Omni-7B} by $11$--$15\%$ and \texttt{LCO-Omni\allowbreak -3B} by $8$--$13\%$; the \omniNano{} variant lands a further $2\%$ below.
\texttt{Omni-Embed-Nemotron-3B} again underperforms at $10.1$ / $9.6\%$ despite being the second-largest model in the table.
Critically, the gap between the \omniSmall{} variant and \texttt{LCO-Omni-7B} is markedly larger on the audio pair ($\sim$$11$--$15\%$ absolute) than on the image pair ($\sim$$6$--$7\%$), under conditions where the centroid distance and paired cosines are within range across all five rows, suggesting the audio bridge into the text-aligned subspace is the weaker of the two projector paths.

\paragraph{Embedding-space geometry across all four modalities.}
Figure~\ref{fig:modality-umap} adds a complementary visual view: a 2D UMAP of $80$ parallel MSR-VTT clips (extracted middle frame, full clip, audio track, and caption), encoded by each model in its natural retrieval protocol.
Three qualitatively distinct patterns appear.
\texttt{LanguageBind} -- the fourth omni-style baseline of Table~\ref{tab:benchmark-summary}, with separate per-modality CLIP-style towers bound to a shared text encoder -- shows the canonical modality-gap pattern from~\cite{modality-gap}: Each modality collapses into a tight, disjoint cluster, with sharp boundaries between them.
\texttt{Omni-Embed-Nemotron-3B} is intermediate. Text is visibly separated from the three other modalities, which themselves overlap somewhat.
The unified-decoder LCO-Omni models and our two frozen-tower \omniSmall{} and \omniNano{} variants produce broadly interleaved geometry: image, video, audio, and text co-mingle in the same neighborhood with substantial overlap, consistent with a single shared embedding space across all four input types rather than four pre-aligned cones.
This geometric picture explains the retrieval-margin behavior reported in the table: Well-mixed clusters have a smaller average modality-cluster offset (lower centroid distance), but their paired-pair coupling also has to compete with same-modality distractions, which is part of why the within-modality R@$1$ gaps remain visible.

\begin{table*}[t]
  \caption{Main benchmark results.
  Bold numeric cells mark the row winner among \omniNano{}, \omniSmall{}, and the strongest open-weight baseline model; bold row labels are benchmark or slice aggregates, and indented rows are task-type averages.
  The ``Strongest open-weight baseline'' column is an orientation point, not a unified controlled ladder.}
  \label{tab:benchmark-summary}
  \centering
  \small
  \setlength{\tabcolsep}{2pt}
  \begin{tabular}{@{}L{0.255\textwidth}rccL{0.245\textwidth}cc@{}}
    \toprule
    Benchmark / task type & \#Tasks & Nano (0.95\,B) & Small (1.57\,B) & Strongest open-weight baseline & Params (B) & Score \\
    \midrule
    \textbf{MIEB Light (Image)} & 51  & 44.24 & 55.01 & LCO-Embedding-Omni-3B & 4.07 & \textbf{61.63} \\
    \quad Image classification & 15 & 44.12 & \textbf{64.12} & LCO-Embedding-Omni-3B & 4.07 & 59.07 \\
    \quad Compositional / vision QA & 11 & 38.85 & 47.61 & LCO-Embedding-Omni-3B & 4.07 & \textbf{52.00} \\
    \quad Image clustering & 2 & 60.46 & \textbf{83.18} & LCO-Embedding-Omni-3B & 4.07 & 73.19 \\
    \quad Visual STS & 4 & 65.92 & \textbf{74.25} & royokong/e5-v & 8.36 & 63.73 \\
    \quad Retrieval & 13 & 25.95 & 31.63 & LCO-Embedding-Omni-3B & 4.07 & \textbf{83.44} \\
    \quad Document retrieval & 6 & 74.21 & \textbf{74.25} & LCO-Embedding-Omni-3B & 4.07 & 72.99 \\
    \midrule
    \textbf{MIEB (Image)} & 125 & 47.87 & 58.00 & siglip-so400m-patch14-384 & 0.88 & \textbf{60.69} \\
    \quad Image classification & 45 & 53.26 & \textbf{68.99} & LCO-Embedding-Omni-3B & 4.07 & 64.30 \\
    \quad Compositional / vision QA & 13 & 40.18 & 49.48 & LCO-Embedding-Omni-3B & 4.07 & \textbf{53.40} \\
    \quad Image clustering & 5 & 72.28 & \textbf{86.01} & LCO-Embedding-Omni-3B & 4.07 & 83.24 \\
    \quad Visual STS & 7 & 75.92 & \textbf{81.74} & LCO-Embedding-Omni-3B & 4.07 & 79.62 \\
    \quad Retrieval & 45 & 30.66 & 37.95 & LCO-Embedding-Omni-3B & 4.07 & \textbf{46.29} \\
    \quad Document retrieval & 10 & 79.25 & 79.25 & Omni-Embed-Nemotron-3B & 4.70 & \textbf{85.64} \\
    \midrule
    \textbf{MIEB Multilingual only (Image)} & 5 & 48.19 & 63.75 & LCO-Embedding-Omni-3B & 4.07 & \textbf{69.04} \\
    \quad Visual STS & 2 & 53.95 & 65.12 & LCO-Embedding-Omni-3B & 4.07 & \textbf{79.62} \\
    \quad Retrieval & 3 & 44.35 & \textbf{62.83} & LCO-Embedding-Omni-3B & 4.07 & 61.99 \\
    \midrule
    \textbf{MMEB-Video (Video)} & 18  & 29.73 & 39.83 & Qwen3-VL-Embedding-8B & 8.14 & \textbf{67.15} \\
    \quad V-CLS (classification) & 5 & 27.85 & 42.73 & Qwen3-VL-Embedding-8B & 8.14 & \textbf{78.39} \\
    \quad V-QA (question answering) & 5 & 39.03 & 44.52 & WeMM-Embedding-8B & 8.77 & \textbf{71.66} \\
    \quad V-RET (retrieval) & 5 & 14.33 & 27.82 & Qwen3-VL-Embedding-8B & 8.14 & \textbf{58.73} \\
    \quad V-MRET (moment retrieval) & 3 & 43.02 & 47.20 & Qwen3-VL-Embedding-8B & 8.14 & \textbf{56.09} \\
    \midrule
    \textbf{MAEB (Audio)} & 30 & 49.69 & 49.96 & LCO-Embedding-Omni-7B & 8.93 & \textbf{52.37} \\
    \quad Retrieval / reranking & 10 & 53.35 & 53.22 & LCO-Embedding-Omni-7B & 8.93 & \textbf{61.67} \\
    \quad Classification / zero-shot & 14 & 53.04 & \textbf{53.71} & LCO-Embedding-Omni-7B & 8.93 & 53.39 \\
    \quad Text matching & 3 & 65.56 & 65.38 & LCO-Embedding-Omni-7B & 8.93 & \textbf{67.30} \\
    \quad Clustering & 3 & 5.96 & 6.13 & clap-htsat-fused & 0.15 & \textbf{22.74} \\
    \bottomrule
  \end{tabular}
  \\[0.4em]
  {\footnotesize MIEB rows use the current MTEB benchmark task composition (MIEB Light 51 tasks, MIEB full 125 tasks, MIEB Multilingual-only 5 tasks). MMEB-Video uses the full $18$-task suite, including MomentSeeker.}
\end{table*}

\section{Ablation Studies}
\label{sec:ablation}
The architecture described in Section~\ref{sec:method} rests on two design choices: which projector layers to train and whether to update an encoder.
This section uses ablation studies to investigate those choices for GELATO.

\subsection{Trainable Parameters}
\label{sec:ablation:trainable-parameters}

The released recipe trains only the second fully connected layer of the vision projector (\texttt{fc\_vision\_2}) and the single linear audio projector (\texttt{fc\_audio}).
This subsection asks whether widening the trainable set --- adding \texttt{fc\_vision\_1}, unfreezing the vision or audio encoder, or running a two-stage continuation --- yields enough improvement to justify the extra parameters and training stages for every task-specific checkpoint.
We sweep five vision configurations (\S\ref{sec:ablation:vision-projector}) and three audio configurations (\S\ref{sec:ablation:audio}).

\paragraph{Setup.}
Runs in this subsection start from \omniSmallRetrieval{}, use global batch $128$ ($32$ per rank $\times$ $4\times$H100), and run for $5\,000$ optimizer steps.
Image ablations use a fast MIEB subset---CIRR-IT2I and NIGHTS-I2I retrieval.
Audio ablations use an 8-task MAEB subset.
For these experiments, the primary trainable projector is randomly initialized at load time: \texttt{fc\_vision\_2} for vision runs and \texttt{fc\_audio} for audio runs.
The remaining layers (encoder, LayerNorm, \texttt{fc\_vision\_1}) retain their pretrained initialization values.

\subsubsection{Vision}
\label{sec:ablation:vision-projector}

We tested which parts of the Qwen3.5 vision stack to train, keeping the rest frozen, evaluating five configurations.
%We sweep five trainable sets for the Qwen3.5 vision path (Figure~\ref{fig:ablation-vision-curves}).
%The Roman numerals below match the figure.

\begin{itemize}
    \item[I] \textbf{\texttt{fc\_vision\_2} only}, lr~$2{\cdot}10^{-4}$ (our configuration).
    \item[II] \textbf{\texttt{fc\_vision\_1} + \texttt{fc\_vision\_2}}, lr~$2{\cdot}10^{-4}$; \texttt{fc\_vision\_1} stays at the Qwen3.5 initialization, \texttt{fc\_vision\_2} is reset.
    \item[III] \textbf{\texttt{fc\_vision\_1} + \texttt{fc\_vision\_2} + vision encoder}, lr~$1{\cdot}10^{-5}$ (dropped $20\times$ because the encoder is unfrozen).
    \item[IV] \textbf{I, then \texttt{fc\_vision\_1} + \texttt{fc\_vision\_2}}, continuing from the stage-I checkpoint.
    \item[V] \textbf{I, then \texttt{fc\_vision\_1} + \texttt{fc\_vision\_2} + vision encoder}, continuing from the stage-I checkpoint.
\end{itemize}

Runs I--III are single-stage ablations from the same reset \texttt{fc\_vision\_2}. Runs IV and V are two-stage continuations that first train run I and then unfreeze additional layers for a second $5\,000$-step stage.

\begin{figure}[t]
\centering
\includegraphics[width=\linewidth]{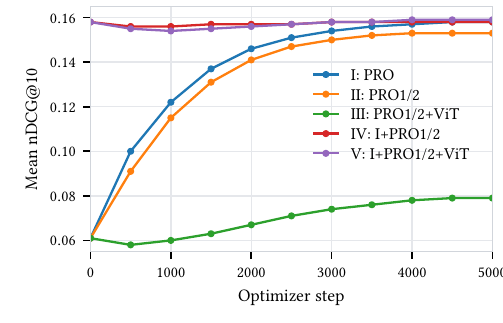}
\caption{Vision ablation tests on CIRR-IT2I and NIGHTS-I2I.
PRO is \texttt{fc\_vision\_2}, PRO1/2 is \texttt{fc\_vision\_1}+\texttt{fc\_vision\_2}, ViT is the vision encoder, and V adds only $0.001$ over I.
\label{fig:ablation-vision-curves}}
\end{figure}

\paragraph{Result:}
Figure~\ref{fig:ablation-vision-curves} displays the results of these tests.
The \texttt{fc\_vision\_2}-only recipe (I) is sufficient: it reaches $0.158$, while training \texttt{fc\_vision\_1} from the start (II) ends slightly lower at $0.153$.
Unfreezing the encoder from step $0$ (III) is clearly harmful, ending at $0.079$.
The two-stage variants test whether I should be followed by a broader continuation stage.
Continuing with \texttt{fc\_vision\_1}+\texttt{fc\_vision\_2} (IV) does not improve the checkpoint, and the broader continuation with the encoder unfrozen (V) reaches only $0.159$, an absolute gain of $0.001$ over I on this 2-task subset.
Run III's collapse to $0.079$ is visible already early in training as a monotone decrease from the shared initial point, indicating that the vision encoder is destabilized as soon as it receives gradients from a randomly initialized \texttt{fc\_vision\_2}; even the $20\times$-reduced learning rate is not enough to prevent the drift when the projector itself has not yet converged.
Runs IV and V remain near the stage-I plateau throughout their continuation, so the small numerical differences among I, IV, and V at step $5\,000$ reflect noise around a basin the projector-only recipe already reached rather than a continued ascent.
That gain is too small to justify a production recipe with an additional continuation stage and extra task-specific adapter/projector artifacts for all four variants of each model size, so the released configuration keeps the simpler frozen-tower choice: train \texttt{fc\_vision\_2} and leave \texttt{fc\_vision\_1}, the vision encoder, and inherited LoRA adapters fixed.

\begin{comment}
\subsection{Vision Encoder}
\label{sec:ablation:vision-encoder}

The Qwen3.5 control in the encoder-swap sweep tests whether retraining a fresh projector without touching the tower can recover the \texttt{fc\_vision\_2}-only configuration.
It matches the run-I band on the 2-task fast-eval subset, so at our training budget we \emph{can} recover our configuration by retraining the projector alone.
The SigLIP2 and CLIP-ViT-L/14 swaps require different modeling wrappers for mismatched image-token shapes and are therefore not part of this controlled sweep.
This is a genuine gap: we have shown that the \emph{projector} is recoverable from scratch, but not that the encoder itself can be swapped without regression.
\end{comment}

\subsubsection{Audio}
\label{sec:ablation:audio}

%We sweep three trainable sets for the Qwen2.5-Omni audio path (Figure~\ref{fig:ablation-audio-freezing}).
We then tested which parts of the Qwen2.5-Omni audio stack to train, keeping the rest frozen, evaluating three configurations.
%The Roman numerals below match the figure.
% The encoder-swap variant 2.b.III keeps the same external projector shape while replacing the tower with Qwen3-Omni's MoE audio encoder and disabling its internal proj1/proj2.

\begin{itemize}
    \item[I] \textbf{\texttt{fc\_audio} only}, lr~$2{\cdot}10^{-4}$ (our configuration).
    \item[II] \textbf{\texttt{fc\_audio} + audio encoder}, lr~$1{\cdot}10^{-5}$; starting from the reset projector.
    \item[III] \textbf{I, then \texttt{fc\_audio} + audio encoder}, continuing from the final I checkpoint, lr~$1{\cdot}10^{-5}$.
%     \item Same Qwen2.5-Omni audio tower, \texttt{fc\_audio} re-initialized.
%     \item Swap the audio tower to Qwen3-Omni's MoE audio encoder.
%         The external projector is $\text{Linear}(1280, d_{\text{text}})$, with proj1/proj2 inside the encoder set to \texttt{Identity} (retaining the merged-projector structure).
\end{itemize}

Runs I and II are single-stage ablations from the same reset \texttt{fc\_audio}. Run III is a two-stage continuation that first trains run I and then unfreezes the audio encoder for a second $5\,000$-step stage.

\begin{figure}[t]
\centering
\includegraphics[width=\linewidth]{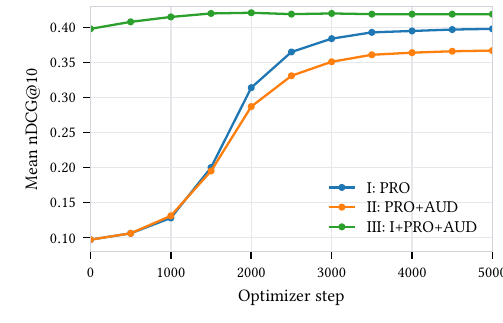}
\caption{Audio ablation tests on UrbanSound8K, CommonVoiceMini21, MACS, GigaSpeech, SpokenSQuAD, Clotho, JamAlt Artist, and JamAlt Lyric.
PRO is \texttt{fc\_audio}, AUD is the audio encoder, and III adds about $0.022$ over I.
\label{fig:ablation-audio-freezing}}
\end{figure}

\begin{comment}
\begin{figure}[t]
\centering
\includegraphics[width=\linewidth]{figures/ablations/section_2b_audio_encoder_swap.pdf}
\caption{Audio encoder swap on the 8-task MAEB subset.
A fresh projector on Qwen2.5-Omni recovers the main path; swapping to Qwen3-Omni regresses sharply.
\label{fig:ablation-audio-encoder-swap}}
\end{figure}
\end{comment}

\paragraph{Result:}
Figure~\ref{fig:ablation-audio-freezing} displays the results of these tests.
The \texttt{fc\_audio}-only recipe (I) is sufficient for this budget: it reaches $0.398$, while unfreezing the audio encoder from step $0$ (II) ends lower at $0.367$.
The two-stage variant tests whether I should be followed by a broader continuation stage.
Continuing with \texttt{fc\_audio}+audio encoder (III) reaches $0.419$, an absolute gain of $0.022$ over I.
Runs I and II diverge within the first $\sim$$1\,000$ steps as the audio encoder begins drifting away from its Qwen2.5-Omni initialization under gradients from a randomly initialized \texttt{fc\_audio}; the curve then plateaus at a lower level than I and does not recover, mirroring the vision-encoder failure mode of run~III in Figure~\ref{fig:ablation-vision-curves}.
Run III by construction tracks run I through step $5\,000$ and only diverges in the continuation stage, where the $+0.022$ lift accumulates gradually rather than as an abrupt jump --- consistent with steady encoder adaptation from an already-converged projector at a $20\times$-reduced learning rate, rather than a one-off alignment correction at the stage boundary.
We therefore keep the released recipe frozen for simplicity, while treating audio-encoder adaptation as a promising future training stage.

This converges with the audio--text geometry analysis (\S\ref{sec:exp:modality-geometry}): the larger LCO baselines maintain a $\sim$$11$--$15\%$ lead over the Small variant on Clotho cross-modal retrieval, and MAEB clustering in Table~\ref{tab:benchmark-summary} shows the Small variant at $6.13$ versus the audio-text specialist \texttt{laion/clap-htsat-fused} ($0.19$B) at $22.74$ --- consistent with the linear \texttt{fc\_audio} bridge discarding intra-modal variance that audio-only K-means relies on.
The single \texttt{fc\_audio} projection is therefore the natural next target for additional trainable parameters, and the $+0.022$ MAEB gain from run III suggests that a budgeted audio-encoder continuation is a promising direction for the next model release.
% The same-encoder control shows that a freshly re-initialized projector on the same encoder converges back to our configuration, mirroring the vision result.
% The Qwen3-Omni swap regresses almost every speech-centric task (CommonVoice $0.870 \to 0.025$, GigaSpeech $0.719 \to 0.001$, SpokenSQuAD $0.388 \to 0.013$) while leaving music-artist retrieval roughly unchanged---unsurprising, since artist retrieval is dominated by the pretrained text encoder matching artist names.
% The drop matches an earlier single-run observation with an alternate projector layout, and we now confirm it under the merged-projector architecture used in this paper: at our training budget, the Qwen3-Omni tower cannot be dropped in without a much longer retraining schedule.

\subsection{Matryoshka Preservation Across Modalities}
\label{sec:ablation:matryoshka}

\begin{figure}[H]
\centering
\includegraphics[width=\linewidth]{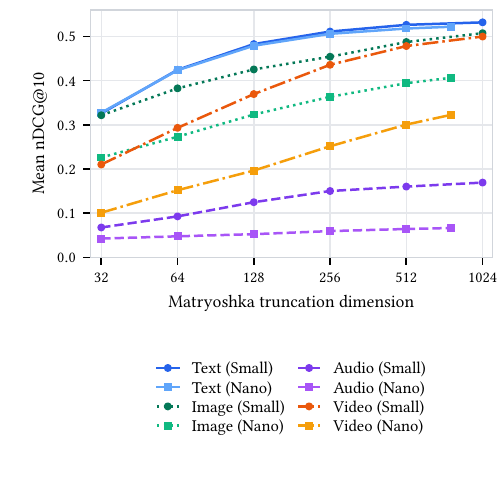}
\caption{Matryoshka prefix tests across modalities.
Curves show mean nDCG@10; line style indicates modality and color shade indicates model size.
\label{fig:ablation-matryoshka}}
\Description{Line chart of mean nDCG@10 versus Matryoshka truncation dimension for text, image, audio, and video retrieval, with small and nano curves for each modality.}
\end{figure}

Figure~\ref{fig:ablation-matryoshka} shows Matryoshka performance under embedding truncation. Image embeddings behave similarly to text ones: both \omniSmall{} and \omniNano{} lose roughly $0.18$--$0.21$ nDCG@10 when truncated to $32$ dimensions. Audio also preserves most of its score at $256$ dimensions, while video degrades much more heavily at small dimensions, indicating weaker Matryoshka preservation for video embeddings.
The curves are near-flat from full dimension down to roughly $128$, with degradation only becoming visible below $64$; the Matryoshka loss therefore appears to keep most retrieval-relevant information in the first nibble of the embedding for image, text, and audio.
The text and image curves overlap almost exactly at every truncation level for a given model size, indicating that the linear \texttt{fc\_vision\_2} projection inherits the prefix structure of the frozen Jina Embeddings v5 Text encoder rather than introducing its own scale-dependent loss.
The video curve diverges from the others well before $64$ dimensions and crosses below half its full-dimension score around $32$, suggesting that aggregating multi-frame information into a single embedding exhausts the residual capacity of the early prefix dimensions faster than the single-frame image case.
The Small and Nano curves for each modality track each other closely under truncation, so the released Matryoshka schedule transfers across both model sizes despite their different embedding widths ($1024$ vs.\ $768$) and projector parameter counts.

% Additional vision and audio encoder swaps require wrapper and processor changes beyond the shared-protocol sweeps above, so they are not listed as rows in the ablation summaries.

\subsection{Training Efficiency}
\label{sec:ablation:efficiency}

This ablation test measures the efficiency gained by GELATO compared to full training.
Table~\ref{tab:ablation-efficiency} shows that projector training makes vision runs $1.8\times$ faster and audio runs $3.2$--$3.9\times$ faster at the $15$k-step budget, with lower peak GPU memory in every case.

\begin{table}[H]
  \caption{Training throughput and peak GPU memory.}
  \label{tab:ablation-efficiency}
  \centering
  \small
  \setlength{\tabcolsep}{2pt}
  \begin{tabular}{@{}L{0.18\linewidth}L{0.18\linewidth}rrrr@{}}
    \toprule
    Setting & Scope & Updated params\textsuperscript{a} & s/step & Peak mem. & $15$k steps \\
    \midrule
    \multicolumn{6}{c}{\omniSmall{}} \\
    \midrule
    Vision & Projector & $4.20$M & $0.413$ & $7.52$\,GiB & $103.3$\,min \\
    Vision & Full & $920.6$M & $0.752$ & $12.96$\,GiB & $188.0$\,min \\
    Audio & Projector & $1.31$M & $0.617$ & $6.06$\,GiB & $154.3$\,min \\
    Audio & Full & $1232.1$M & $1.989$ & $19.53$\,GiB & $497.3$\,min \\
    \midrule
    \multicolumn{6}{c}{\omniNano{}} \\
    \midrule
    Vision & Projector & $2.36$M & $0.181$ & $6.94$\,GiB & $45.2$\,min \\
    Vision & Full & $311.6$M & $0.329$ & $10.02$\,GiB & $82.3$\,min \\
    Audio & Projector & $0.98$M & $0.447$ & $5.77$\,GiB & $111.7$\,min \\
    Audio & Full & $847.8$M & $1.764$ & $16.08$\,GiB & $440.9$\,min \\
    \bottomrule
  \end{tabular}
  \\[0.4em]
  \begin{minipage}{0.97\linewidth}
  \footnotesize
  \textsuperscript{a}\,Parameters that actually receive updates. In projector scope these are the trained projector weights (\texttt{fc\_vision\_2} or \texttt{fc\_audio}) plus the modality-delimiter token rows; the token-embedding matrix is held in the optimizer with gradients masked to those rows, so its remaining rows do not change but its optimizer state is included in peak memory.
  \end{minipage}
\end{table}

\section{Conclusion}
\label{sec:conclusion}

We introduce GELATO, a novel approach to constructing multimodal embedding models by connecting frozen pre-trained modality-specific encoders directly to a frozen text embedding model via compact and easily trained projectors.
The result of this research, the \omniBase{} model suite, is also presented. 
These models add vision and audio to the Jina Embeddings v5 Text models, yielding a competitive set of models for broad cross-modality applications. 
Using GELATO, text-only embedding models that were never trained on vision or audio can be extended to photos, documents, video, speech, music, and sounds by training a single projector layer per modality while preserving text-only performance.

\omniSmall{} is the best-performing open-weight embedding model below $2$B parameters that supports text, audio, images, and video.
Against a baseline of comparable models, including modality-specific and VLM-derived embedders, it is particularly strong on visual document retrieval.
\omniSmall{} and \omniNano{} extend completely different text embedding models with different backbone architectures, suggesting that GELATO is an extensible strategy with broad application, outside of the \omniBase{} suite and for additional modalities.
This is a potential subject for future research.

The ablations suggest that projector-only alignment can serve as a compatibility-preserving initialization for rich multimodal training.
Future work will investigate the choice of non-text encoders, which is inadequately explored in this paper.
Furthermore, an investigation of training options under different conditions is indicated, like jointly training projectors for multiple modalities together.
We also note that \omniBase{} is comparatively closer to the baselines on moment retrieval than on the other video sub-tasks, but overall video performance remains weak.
We hope to improve performance in this area in future models.

\bibliographystyle{ACM-Reference-Format}
\bibliography{references}

\end{document}